\documentclass{article}
\usepackage{spconf,amsmath,amssymb,graphicx,hyperref}
\usepackage{booktabs}
\usepackage{makecell}
\usepackage{multirow}
\usepackage{threeparttable}
\usepackage{tabularx}
\usepackage{xcolor}
\usepackage{subcaption}
\usepackage{mathabx}

\title{Adaptive Fisher-Whitened Cross-Covariance for Low-Resource Speech Recognition}

\name{Asmee Mishra$^1$, Mengjie Qian$^1$, Brechtje Post$^2$, Kate Knill$^1$}

\address{
$^1$Department of Engineering, University of Cambridge, United Kingdom\\
$^2$Phonetics Laboratory, University of Cambridge, United Kingdom
}

\begin{document}

\ninept
\maketitle

\begin{abstract}
Adapting multilingual speech foundation models to low-resource languages remains difficult, especially for languages that are poorly represented during pre-training.
While parameter-efficient fine-tuning (PEFT) reduces the cost of adapting large models, conventional approaches such as LoRA rely on generic low-rank parameterizations and do not explicitly use downstream task information to define the adaptation subspace. To investigate whether task-informed PEFT can better support low-resource ASR, we apply Fisher-Whitened Cross-Covariance Analysis (FCCA) to Whisper and Qwen3-ASR, and introduce two complementary extensions: Asymmetric-Coupled FCCA (AC-FCCA), which exploits structured cross-layer sharing, and Adaptive-Rank FCCA (AR-FCCA), which reallocates adaptation capacity across projection matrices under a fixed parameter budget. Under controlled multilingual experiments, we evaluate these approaches on languages that are poorly represented or unsupported during pre-training alongside well-represented languages. Standard FCCA is competitive with, and usually outperforms, trainable-parameter-budget-matched LoRA. AR-FCCA provides the most consistent improvement over standard FCCA across both model architectures, with statistically significant gains in several evaluation settings, while retaining the same number of trainable parameters. These results show that task-informed subspace construction can be effective for low-resource speech adaptation, and that adaptive rank allocation provides a robust way to improve parameter efficiency without increasing model capacity.
\end{abstract}


\section{Introduction}
\label{sec:intro}

Speech foundation models, such as Whisper~\cite{radford2022robust} and Qwen3-ASR~\cite{shi2026qwen3}, have substantially improved multilingual automatic speech recognition (ASR), allowing a single model to recognise dozens of languages. However, many languages remain underrepresented during pre-training, including low-resource, endangered and regional languages, for which only a few hours of labelled speech are available. Developing speech technologies for these languages benefits practical applications, linguistic research and language documentation~\cite{lonergan2022automatic,lonergan2025fotheidil}. In particular, speech technologies can accelerate the documentation, annotation, and analysis of endangered languages, providing scalable tools for phonetic, phonological, and sociolinguistic studies. Consequently, efficiently adapting speech foundation models with limited labelled data has become an important research problem for multilingual low-resource ASR. 

Full fine-tuning remains an effective adaptation strategy but requires updating all model parameters, resulting in substantial computational and memory costs. Parameter-efficient fine-tuning (PEFT) addresses this limitation by adapting only a small subset of parameters while maintaining competitive performance. Representative approaches include adapter tuning \cite{houlsby2019parameter,pfeiffer2020adapterhub}, prefix tuning \cite{liliang2021prefix}, prompt tuning \cite{lester2021prompt}, and Low-Rank Adaptation (LoRA) \cite{hu2022lora}, with LoRA becoming one of the dominant approaches for adapting speech foundation models. 
Recent studies have further investigated continual language learning~\cite{qian2024learn} and source language adaptation for low-resource ASR~\cite{dang2026sequential}. Recent PEFT methods have increasingly exploited downstream data to identify more informative adaptation directions~\cite{yang2024corda,li2025aira,das2025consnotrainlora,li2025nora}, for example through gradient-informed initialisation in LoRA-GA~\cite{wang2024loraga} and activation-based subspace selection and rank allocation in EVA~\cite{paischer2024eva}. Fisher-Whitened Cross-Covariance Analysis (FCCA) takes this further by deriving paired low-rank adaptation subspaces from Fisher-whitened input--error statistics~\cite{ye2026fcca}. This task-informed PEFT has shown promising results for large language models (LLMs), but its effectiveness for speech foundation models, particularly under low-resource adaptation, remains unexplored.

This work investigates whether task-informed PEFT transfers effectively to multilingual low-resource ASR by adapting FCCA to speech foundation models. The original FCCA study focuses on decoder-only autoregressive LLMs (i.e. Qwen2.5-3B-Instruct~\cite{qwen2.5}), whereas speech models such as Whisper use an encoder--decoder architecture that must first learn representations from continuous acoustic input. These differences make it unclear whether the same task-informed subspace construction will behave similarly for speech recognition. Building upon FCCA, we propose two complementary approaches for improving task-informed adaptation. \textbf{Asymmetric-Coupled FCCA (AC-FCCA)} investigates whether structured cross-layer parameter sharing can improve adaptation efficiency, while \textbf{Adaptive-Rank FCCA (AR-FCCA)} redistributes adaptation capacity across projection matrices under a fixed parameter budget. These two approaches investigate different aspects of task-informed PEFT: how adaptation subspaces can be shared across layers, and how adaptation capacity can be distributed across matrices.

In this work, we evaluate the proposed methods on low-resource ASR across multiple languages using Whisper and Qwen3-ASR, chosen to cover different model architectures and their strong performance on multilingual ASR. Experimental results demonstrate that AR-FCCA consistently improves standard FCCA across diverse multilingual benchmarks while remaining competitive with strong PEFT baselines. The main contributions of this work are:
(1) the first systematic investigation of task-informed PEFT (i.e. FCCA) for multilingual low-resource ASR;
(2) proposing two complementary approaches, AC-FCCA and AR-FCCA, extending FCCA through complementary cross-layer sharing and adaptive-rank allocation strategies; and
(3) extensive experiments across multiple languages and two speech foundation models.

\section{Proposed Methods}
\label{sec:methods}
Ultra-low-parameter PEFT is attractive because it reduces trainable and optimiser-state memory, but at such small budgets the quality of the chosen update subspace becomes critical. Several LoRA variants use task-dependent statistics to improve the initial adaptation subspace. LoRA-GA~\cite{wang2024loraga} uses downstream gradients, EVA~\cite{paischer2024eva} uses activation statistics, and CorDA~\cite{yang2024corda} uses task-conditioned activation covariance to orient a decomposition of the pretrained weights, among other approaches~\cite{li2025aira,das2025consnotrainlora,li2025nora}. However, these retain LoRA-sized trainable factors. FCCA \cite{ye2026fcca} instead fixes task-informed left and right bases and trains only a small core matrix, yielding a substantially smaller adaptation budget. We first review FCCA and then introduce two complementary extensions: AC-FCCA for structured cross-layer sharing and AR-FCCA for adaptive rank allocation. 

\subsection{Fisher-Whitened Cross-Covariance Adaptation (FCCA)}
\label{subsec:fcca}
For an adapted weight matrix $W\in\mathbb{R}^{m\times n}$, FCCA parameterises the update as
\begin{equation}
    W' = W + sPRQ^\top ,
\end{equation}
where $P\in\mathbb{R}^{m\times r}$ and $Q\in\mathbb{R}^{n\times r}$ are fixed task-informed bases and only $R\in\mathbb{R}^{r\times r}$ is trainable.

Given layer input $x$ and back-propagated output error $\delta$, FCCA estimates
\[
    G=\mathbb{E}[\delta x^\top], \hspace{2pt}
    A=\operatorname{diag}(\mathbb{E}[x^{\odot2}])+\epsilon I, \hspace{2pt}
    D=\operatorname{diag}(\mathbb{E}[\delta^{\odot2}])+\epsilon I,
\]
and forms the Fisher-whitened cross-covariance
$\widetilde{G}=D^{-1/2}GA^{-1/2}$.
With the rank-$r$ decomposition $\widetilde{G}\approx U_r\Sigma_rV_r^\top$, the bases are obtained by inverse whitening,
$P_0=D^{-1/2}U_r, Q_0=A^{-1/2}V_r$,
followed by thin QR orthogonalisation. 

Unlike LoRA, whose trainable parameter count scales with $r(m+n)$, FCCA trains only the $r^2$ parameters in $R$, where $r \ll m,n$. This frozen-core formulation is similar to LoRA-XS~\cite{balazy2024loraxs}, but differs in how the fixed bases are constructed: LoRA-XS uses singular directions from the pretrained weight spectrum, whereas FCCA derives task-informed directions from downstream statistics. On LLM benchmarks, FCCA has shown stronger matched-budget performance than LoRA-XS-based alternatives~\cite{ye2026fcca}.


\subsection{Asymmetric-Coupled FCCA (AC-FCCA)}
\label{subsec:ac-fcca}

Standard FCCA constructs independent left and right adaptation subspaces for each matrix, although neighbouring Transformer layers may share task-relevant structure. 
We quantify cross-layer alignment between orthonormal rank-$r$ bases $X$ and $Y$ by
\[
S(X,Y)=r^{-1}\|X^\top Y\|_F^2,
\]
with larger values indicating greater shared subspace structure. 

Our preliminary analysis shows projection-dependent asymmetry: \(q/k/v\) projections align more strongly across layers on the input side, whereas \(o\) aligns more strongly on the output side, with alignment strongest between adjacent layers. AC-FCCA therefore chooses to couple only this better-aligned side across adjacent layers while keeping the opposite side layer-specific. 

\textbf{Asymmetric one-sided FCCA (AC-FCCA-H).}
For each adjacent-layer pair $(\ell,\ell+1)$, we share a rank-$r$ basis on the better-aligned side. For \(q/k/v\),
\[
V=\operatorname{TopEig}_{r}\!\left(\sum_{j\in\{\ell,\ell+1\}}\widetilde G_j^\top\widetilde G_j\right),
\qquad
U_j=\operatorname{orth}(\widetilde G_jV),
\]
whereas for \(o\),
\[
U=\operatorname{TopEig}_{r}\!\left(\sum_{j\in\{\ell,\ell+1\}}\widetilde G_j\widetilde G_j^\top\right),
\qquad
V_j=\operatorname{orth}(\widetilde G_j^\top U).
\]
The shared basis is defined in Fisher-whitened coordinates.
After layer-specific FCCA unwhitening and QR orthogonalisation,
each matrix obtains native bases $P_j,Q_j$ and
$\Delta W_j=P_jR_jQ_j^\top$.

\textbf{Shared--private asymmetric FCCA (AC-FCCA-SP).}
To relax full one-sided sharing, the coupled rank-$128$ subspace is split into $r_s=64$ shared and $r_p=64$ private directions. 

Define
\[
C_j=
\begin{cases}
\widetilde G_j^\top\widetilde G_j, & q/k/v,\\
\widetilde G_j\widetilde G_j^\top, & o.
\end{cases}
\]
For each adjacent-layer pair, shared basis $S$ and private bases $P_j$ maximise
\begin{equation}
J=\sum_j\left[
\operatorname{tr}(S^\top C_jS)
+\operatorname{tr}(P_j^\top C_jP_j)
\right],
\end{equation}
subject to
$S^\top S=P_j^\top P_j=I$ and $S^\top P_j=0$.
We optimise this objective by block-coordinate ascent, alternating
\[
P_j=\operatorname{TopEig}_{r_p}
\!\left(\Pi_S^\perp C_j\Pi_S^\perp\right),
\qquad
\Pi_S^\perp=I-SS^\top,
\]
with
\[
S=\operatorname{TopEig}_{r_s}
\!\left(
\Pi_{P_{\cup}}^\perp
\left(\sum_j C_j\right)
\Pi_{P_{\cup}}^\perp
\right),
\]
where $P_{\cup}$ is the numerical union span of the private bases. Updates alternate to convergence; the shared and private bases are then concatenated, and the opposite-side basis is recovered as in AC-FCCA-H.

\subsection{Adaptive-Rank FCCA (AR-FCCA)}
\label{subsec:ar-fcca}
Recent adaptive-rank LoRA methods relax the assumption of a uniform
rank across weight matrices, instead allocating the rank budget
according to matrix importance~\cite{zhang2023adalora,ilora2026}. Motivated by this idea, we extend FCCA with matrix-specific ranks while preserving the same total trainable-core budget. FCCA naturally provides a task-informed criterion for rank allocation. For matrix $\ell$, let
$\widetilde{G}_\ell = U_\ell \Sigma_\ell V_\ell^\top$.
Under the FCCA local quadratic approximation, the utility of the optimal rank-$r$ update is proportional to the retained Fisher-whitened spectral energy $E_\ell(r) = \sum_{k=1}^{r}\sigma_{\ell,k}^{2}$,
where $\sigma_{\ell,k}$ is the $k$-th singular value of
$\widetilde{G}_\ell$.
Since the trainable FCCA core for matrix $\ell$ contains $r_\ell^2$ parameters, AR-FCCA selects matrix-specific ranks under a fixed global parameter budget:
\begin{equation}
    \max_{\{r_\ell\}}   \sum_{\ell=1}^{L}E_\ell(r_\ell) 
    \quad
    \mathrm{s.t.}
    \quad
    \sum_{\ell=1}^{L}r_\ell^2=B,
    \qquad
    B=Lr_0^2 ,
\end{equation}
where $r_0$ is the uniform FCCA baseline rank. This reallocates adaptation capacity toward matrices that retain more task-relevant spectral energy per additional trainable parameter, while keeping the overall core budget unchanged.
At the selected rank, the standard FCCA construction is applied:
$
P_\ell = \operatorname{qr}(D_\ell^{-1/2}U_{\ell,:r_\ell}), \qquad
Q_\ell = \operatorname{qr}(A_\ell^{-1/2}V_{\ell,:r_\ell})$.
The bases remain frozen and only $R_\ell\in\mathbb{R}^{r_\ell\times r_\ell}$ is trained, giving
$\Delta W_\ell = P_\ell R_\ell Q_\ell^\top$.
In all experiments, $r_0=128$, and AR-FCCA exactly matches the total trainable-core budget of uniform FCCA.

\section{Experimental Setup}
\label{sec:exp_setup}

\textbf{Datasets.} Experiments use five FLEURS languages~\cite{conneau2023fleurs}, spanning low-resource targets and better-represented multilingual controls. Asturian, Sorani Kurdish, and Kyrgyz form the primary low-resource evaluation, while Mandarin and Persian test whether the observed behaviour extends to better-represented languages. For Whisper, unseen languages use tokens from closely related languages. All adaptation uses only the target-language FLEURS training split (7.5--10.5 h), with hyperparameters selected on validation and final results reported on the held-out test set.

\begin{table}[!htbp]
    \centering
    \caption{FLEURS dataset statistics (hours) for target languages.}
    \vspace{-3mm}
    \label{tab:fleurs_stats}
    \begin{tabular}{@{ }l|cccc@{ }}
        \toprule
        Language & Lang. Token & Train  & Val & Test \\
        \midrule
        Asturian & $\langle|\mathrm{es}|\rangle$ & 7.5  & 0.9 & 2.4 \\
        Sorani Kurdish & $\langle|\mathrm{fa}|\rangle$ & 10.5 & 1.2 & 3.0 \\
        Kyrgyz & $\langle|\mathrm{kk}|\rangle$ & 9.3 & 1.3 & 3.2 \\
        Mandarin & $\langle|\mathrm{zh}|\rangle$ & 9.7 & 1.3 & 3.1 \\
        Persian & $\langle|\mathrm{fa}|\rangle$ & 10.0 & 1.5 & 3.7 \\
        \bottomrule
    \end{tabular}
\end{table}

\textbf{Models.} Experiments use Whisper medium~\cite{radford2022robust} and Qwen3-ASR-1.7B~\cite{shi2026qwen3}. Whisper medium is a 769M-parameter encoder--decoder model with cross-attention, whereas Qwen3-ASR combines an audio encoder with an autoregressive LLM-style decoder. Whisper is evaluated on all five languages; Qwen3-ASR provides cross-model validation on Asturian and Sorani Kurdish, neither of which is among its supported languages.

\textbf{Baselines and compared approaches.} The following methods are used as baselines in the experiments: \textbf{Vanilla}, using the pretrained model directly without any fine-tuning; \textbf{full fine-tuning (FFT)}, updating all model parameters; \textbf{LoRA}~\cite{hu2022lora}, a widely-used PEFT baseline.
These are compared with standard \textbf{FCCA}~\cite{ye2026fcca} and the proposed \textbf{AC-FCCA} and \textbf{AR-FCCA} variants.

\textbf{Adaptation Configurations.}
For Whisper, LoRA and FCCA-based methods adapt the \(q/k/v/o\) projections in encoder self-attention, decoder self-attention, and decoder cross-attention (288 matrices); for Qwen3-ASR, the corresponding audio-encoder and text-decoder projections are adapted (208 matrices). FCCA and its variants use 256 calibration examples sampled from the training set with the same seed, and base rank \(r=128\), training 0.61\% and 0.20\% of Whisper-medium and Qwen3-ASR parameters, respectively. Calibration is a one-off cost, requiring approximately 20, 40, 90, and 28\,s for FCCA, AC-FCCA-H, AC-FCCA-SP, and AR-FCCA, respectively, on an RTX 6000 GPU.

\textbf{Training Configuration.}
Hyperparameters are selected by a small sequential validation-set search. For Whisper-medium, FFT searches learning rates in \([4,7]\times10^{-5}\) with effective batch size (EBS) 32, LoRA in \([1,4]\times10^{-4}\) with EBS 32, and FCCA in \([1,15]\times10^{-4}\) with EBS 8. FCCA uses \(r=128\), while LoRA uses \(r=128\), \(r=8\), and \(r=6\). All systems use AdamW, weight decay 0.01, linear warm-up ratio 0.05, gradient clipping at 1.0, bf16 precision, and at most 12 epochs with early stopping. After selecting the best standard FCCA configuration, its optimisation hyperparameters are reused unchanged for AC-FCCA-H, AC-FCCA-SP, and AR-FCCA, avoiding variant-specific tuning.

\textbf{Decoding Configuration.}
All systems use deterministic greedy decoding (beam size 1) without an external language model. The target language and transcription task are specified at inference, and the repetition guard described below is applied during generation.

\textbf{Post-processing and Evaluation.}
Before scoring, references and hypotheses undergo NFKC normalization~\cite{unicode_nfkc}, case-folding, punctuation/symbol and control-character removal, Unicode-digit canonicalization, and whitespace collapse. Persian additionally canonicalizes Arabic-script variants for evaluation. Mandarin hypotheses are converted from Traditional to Simplified Chinese with OpenCC~\cite{opencc}; whitespace is then removed from references and hypotheses before CER computation. We report CER for Mandarin and WER for all other languages. To suppress decoding hallucinations, generation terminates when any token cycle of length 1--8 repeats six consecutive times.

\section{Results and Discussion}
\label{sec:results}
\subsection{Whisper Baseline Performance}
We first present the baseline experiments on Whisper under different adaptation strategies. As shown in Table~\ref{tab:main_results}, Vanilla Whisper shows low CER on Mandarin and reasonable WER on Persian, both of which are supported languages in Whisper, but degrades substantially on unsupported languages. This contrast highlights the  difficulty of recognising languages with limited pre-training support. 
Full fine-tuning provides the strongest conventional adaptation baseline across most languages. On Mandarin, however, Vanilla Whisper already achieves 8.88\% CER, leaving limited room for improvement with fewer than 10 hours of training data. LoRA with rank 128 (LoRA-128) approaches FFT while updating substantially fewer parameters. To separate the effect of adaptation strategy from trainable-parameter budget, we additionally include rank-8 LoRA (LoRA-8), which has a parameter budget comparable to the FCCA-based methods. Compared to LoRA-128, LoRA-8 performs 2--8\% WER worse, which is expected given its smaller update space.

\subsection{Task-informed PEFT - FCCA variants}
Here, we investigate the performance of standard FCCA (Section~\ref{subsec:fcca}) and proposed FCCA variants (Section~\ref{subsec:ac-fcca} and \ref{subsec:ar-fcca}) on Whisper.
We use rank 128 for FCCA to match the rank of LoRA-128, which provides performance close to FFT while remaining parameter efficient. Owing to the frozen-core formulation, rank-128 FCCA trains only 0.61\% of Whisper parameters, giving a trainable-parameter budget comparable to LoRA-8. This allows two complementary comparisons: LoRA-128 provides a rank-matched baseline, while LoRA-8 provides a parameter-matched baseline.

FCCA trains 16 times fewer parameters than LoRA-128 and yields WER 0.36–7.14 percentage points higher than LoRA-128. Moreover, under the parameter-matched comparison, FCCA outperforms LoRA-8 on 3 languages in Whisper and almost matches it in others, suggesting that task-informed subspace construction is more effective than the budget-matched standard LoRA method. FCCA and all FCCA variants on Asturian and Kyrgyz, as well as AC-FCCA-H and AR-FCCA on Mandarin, give significant results at $p<0.05$ under MAPSSWE against LoRA-8.

The AC-FCCA variants introduce structured cross-layer sharing on top of standard FCCA. They outperform FCCA across four of the five Whisper languages and AC-FCCA-H yields significant results at $p<0.05$ under MAPSSWE on two languages, improving WER from 44.97\% to 44.35\% on Sorani Kurdish and CER from 10.46\% to 10.19\% on Mandarin. Our preliminary analysis had shown that both AC-FCCA constructions achieved higher Fisher-weighted gradient capture than standard FCCA on a disjoint probe set, suggesting improved preservation of task-relevant update structure.

AR-FCCA gives the most consistent improvement over standard FCCA; it reduces WER by 0.32–1.13 percentage points across all languages and achieves statistically significant gains at $p<0.05$ under MAPSSWE over standard FCCA on Sorani Kurdish and Mandarin. The improvements are observed across both unsupported and supported languages, showing that adaptive rank allocation provides a robust extension to standard FCCA without increasing the overall parameter budget.

\begin{table}[t]
    \centering
    \caption{WER (\%) on FLEURS test sets using Whisper medium; CER (\%) for Mandarin.}
    \label{tab:main_results}
    \vspace{-3mm}

    \small
    \setlength{\tabcolsep}{2.5pt}

    \begin{tabular}{@{}l|c|ccc|cc@{}}
        \toprule
        & & \multicolumn{3}{c|}{Unsupported} & \multicolumn{2}{c}{Supported} \\
        Method & \%Para & Ast. & Sor. & Kyr. & Man. & Per. \\
        \midrule
        Vanilla & 0 & 51.11 & 114.03 & 90.46 & 8.88 & 47.30 \\
        FFT & 100 & 15.60 & 35.53 & 18.39 & 9.97 & 14.20 \\
        LoRA-128 & 9.82 & 15.96 & 37.83 & 19.75 & 10.10 & 14.92 \\
        LoRA-8 & 0.61 & 19.17 & 44.29 & 25.63 & 10.66 & 16.16 \\
        \midrule
        FCCA & \multirow{4}{*}{0.61} & 17.07$^\ast$ & 44.97 & 22.25$^\ast$ & 10.46 & 16.20 \\
        AC-FCCA-H & & 16.97$^\ast$ & 44.35$^\dagger$ & 22.12$^\ast$ & \textbf{10.19}$^{\dagger\ast}$ & 16.65 \\
        AC-FCCA-SP & & \textbf{16.64}$^\ast$ & 44.83 & \textbf{21.85}$^\ast$ & 10.41 & 16.33 \\
        AR-FCCA & & 16.67$^\ast$ & \textbf{43.84}$^\dagger$ & 21.93$^\ast$ & 10.37$^{\dagger\ast}$ & \textbf{15.74} \\
        \bottomrule
    \end{tabular}

    \vspace{0.5mm}
    {\scriptsize
    $^\ast$ denotes $p<0.05$ versus LoRA-8;$^\dagger$ denotes $p<0.05$ versus FCCA.
    }
\end{table}

\subsection{Generalization to Qwen3-ASR}
This subsection investigates whether the observations on Whisper generalize to a structurally different speech foundation model. Qwen3-ASR is evaluated on Asturian and Sorani Kurdish, both of which are unsupported by the pretrained model. As shown in Table~\ref{tab:qwen_results}, Vanilla Qwen3-ASR performs poorly on both languages, with WERs of 48.94\% and 105.30\%, respectively. As with Whisper, FFT and LoRA-128 give largest improvement on WER; for Asturian, FFT and LoRA are almost matched at 16.06\% and 16.05\%.

Standard FCCA substantially improves the pretrained model, reducing WER to 19.61\% on Asturian and 44.22\% on Sorani, but as with Whisper, it still trails the rank-matched LoRA-128. FCCA also beats its trainable-parameter-matched LoRA version (LoRA-6) by 0.17\% WER on Asturian and 0.77\% on Sorani Kuridish. However, the AC-FCCA variants which offered improvements on FCCA in Whisper, increase WER in Qwen3-ASR. In contrast, AR-FCCA again gives the strongest FCCA-based performance, reducing WER further to 18.56\% and 42.98\%, respectively. These improvements over standard FCCA are statistically significant for both Asturian ($p=0.001$) and Sorani ($p=0.05$) under MAPSSWE.

\begin{table}[!t]
    \centering
    \caption{WER (\%) on FLEURS test sets using Qwen3-ASR-1.7B.}
    \label{tab:qwen_results}
    \vspace{-3mm}
    \begin{tabular}{l|c|cc}
        \toprule
        Method & \%Para & Ast. & Sor. \\
        \midrule
        Vanilla & 0 & 48.94 & 105.30 \\
        FFT & 100 & 16.06 & 37.93 \\
        LoRA-128 &  4.50 & 16.05 & 40.78 \\
        LoRA-6 & 0.21 & 19.78 & 44.99 \\
        \midrule
        FCCA & \multirow{4}{*}{0.20} & 19.61 & 44.22 \\
        AC-FCCA-H & & 19.63 & 44.36 \\
        AC-FCCA-SP & & 19.94 & 44.64 \\
        AR-FCCA & & \textbf{18.56}$^\dagger$$^\ast$ & \textbf{42.98}$^\dagger$$^\ast$ \\
        \bottomrule
    \end{tabular}
\vspace{1pt}
\begin{minipage}{0.94\columnwidth}
    \scriptsize
    \setlength{\baselineskip}{7pt}
   $^\ast$ denotes $p<0.05$ versus LoRA-6; $\dagger$ denotes $p<0.05$ versus FCCA.
\end{minipage}
\end{table}
\vspace{-3mm}

\subsection{Analysis of Adaptive Rank Allocation}
\begin{figure}[t]
    \centering
    \includegraphics[trim=0 15mm 0 10mm,width=0.9\columnwidth]{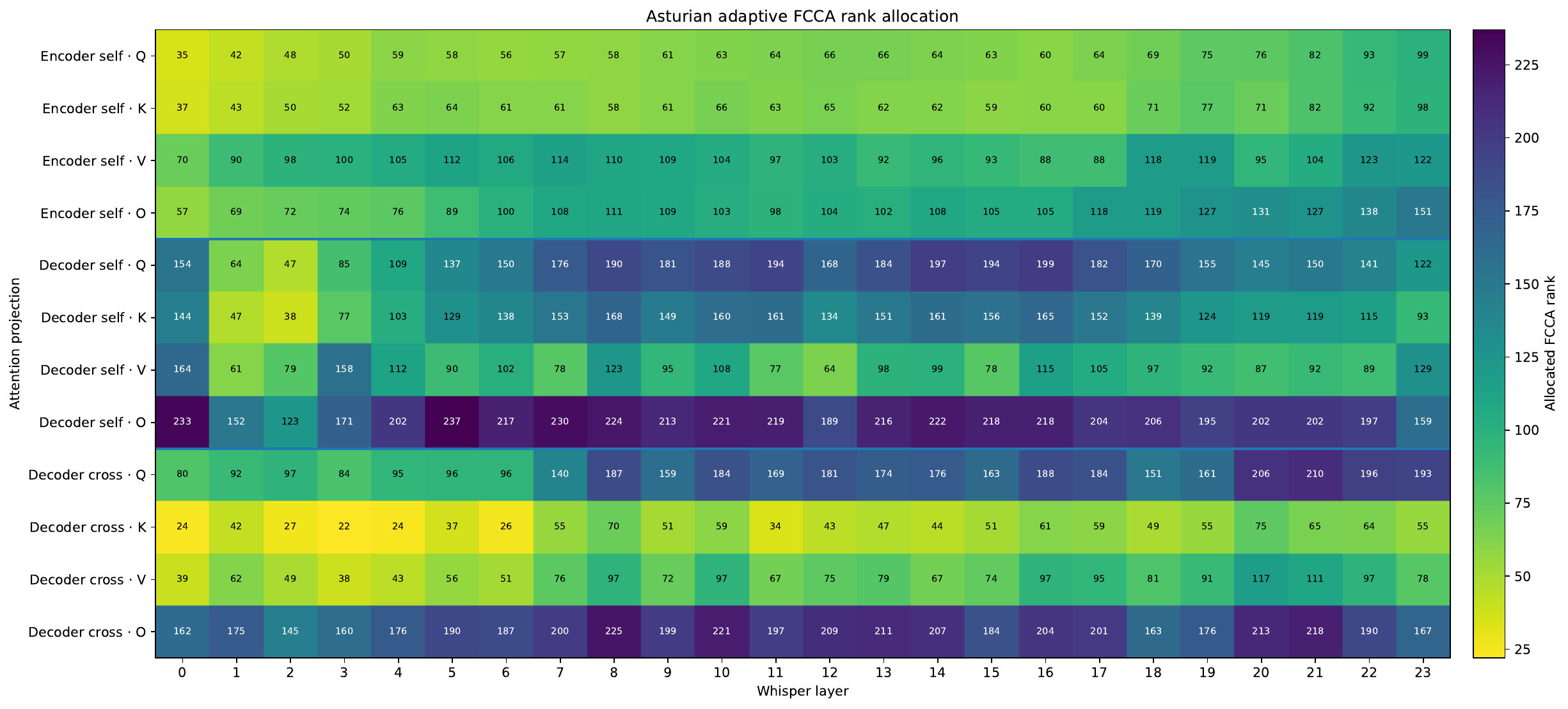}
    \caption{Adaptive FCCA rank allocation across Whisper model layers and attention projections for Asturian. Darker cells indicate larger allocated rank.}
    \label{fig:rank_allocation}
\end{figure}

The consistent gains from AR-FCCA across Whisper and Qwen3-ASR motivate a closer examination of this approach. Figure~\ref{fig:rank_allocation} shows the rank allocation for Whisper on Asturian. Rank allocation patterns are highly consistent across languages within each model. AR-FCCA jointly chooses per-matrix ranks to maximize retained Fisher-whitened spectral energy under a fixed global budget, so larger ranks are assigned where adding further directions continues to capture substantial task-relevant spectral energy. In Whisper, this concentrates rank in decoder output projections, while Qwen3-ASR assigns high ranks to \(q/o\) in the text tower but to \(v/k\) in the audio tower, with audio rank also tending to increase with depth.

To examine why AR-FCCA consistently improves over uniform-rank FCCA, we further test whether the rank allocation selected before training remains advantageous as optimisation progresses. For each intermediate checkpoint, the FCCA statistics and Fisher-whitened singular spectra are recomputed, while keeping the AR-FCCA ranks $r_\ell^{\mathrm{AR}}$ fixed to their initial values. We then compare this fixed allocation with uniform rank $128$ under the same total trainable-core budget.
Following the FCCA local quadratic formulation, the utility of a rank allocation is measured by the retained spectral energy
$\sum_{\ell}\sum_{k\le r_\ell}\sigma_{\ell,k}^{2}$,
which is proportional to the best local rank-constrained loss reduction under the FCCA surrogate. At checkpoint $e$, we define
$\Delta_{\mathrm{fresh}}^{(e)}=100\left(\frac{\sum_{\ell}\sum_{k=1}^{r_\ell^{\mathrm{AR}}}(\sigma_{\ell,k}^{(e)})^2}{\sum_{\ell}\sum_{k=1}^{128}(\sigma_{\ell,k}^{(e)})^2}-1\right)\%.
$
Across both Whisper and Qwen3-ASR and all evaluated languages, $\Delta_{\mathrm{fresh}}^{(e)}>0$ at every checkpoint. This indicates that the rank allocation selected from the initial FCCA geometry continues to retain more task-relevant spectral utility than uniform rank allocation throughout training.


\section{Conclusion}
This work investigated task-informed PEFT for multilingual low-resource ASR through FCCA and two complementary extensions. Across Whisper and Qwen3-ASR, standard FCCA provided a highly parameter-efficient alternative to conventional PEFT with matched parameter budget, while the AC-FCCA variants gave only limited additional gains. AR-FCCA provided the most consistent improvements over standard FCCA by reallocating the same trainable budget across projection matrices, with gains observed across both speech foundation models and multiple languages and statistically significant improvements in several settings. Further analysis showed that AR-FCCA assigns markedly different ranks across layers and projections, and that these allocations retain higher FCCA spectral utility than uniform rank allocation throughout training. A limitation is the lack of systematic parameter-matched LoRA comparisons such as LoRA-XS, which we leave for future work.

\section*{Acknowledgements}
This paper reports on research supported by Cambridge Language Sciences Incubator Fund.


\bibliographystyle{IEEEbib}
\bibliography{refs,strings}

\end{document}